\documentclass[conference]{IEEEtran}
\usepackage[utf8]{inputenc}
\usepackage{cite}
\usepackage{tikz}
\usepackage{acro}
\usetikzlibrary{mindmap}
\usepackage{graphicx}
\usepackage{subcaption}
\usepackage{adjustbox}
\usepackage{stfloats} 
\usepackage{flushend} 
\usepackage{caption}
\usepackage{url}
\usepackage[numbers,sort&compress]{natbib} 
\usepackage{graphicx}
\usepackage{pdflscape}
\usepackage{adjustbox}
\usepackage{booktabs}
\usepackage{tabularx}
\usepackage{rotating}
\usepackage{afterpage}
\usepackage{lipsum}
\usepackage{amsmath} 
\usepackage{amssymb} 
\usepackage{hyperref}
\usepackage{cleveref} 
\usepackage{pifont}
\newcommand{\xmark}{\ding{55}}%
\usepackage{tikz}
\usetikzlibrary{arrows.meta, positioning}
\usepackage{xcolor}
\DeclareAcronym{3D}{
    short = 3D,
    long  = three-dimensional
}
\DeclareAcronym{API}{
    short = API,
    long  = application programming interface
}
\DeclareAcronym{AUV}{
    short = AUV,
    long  = autonomous underwater vehicle
}
\DeclareAcronym{BT}{
    short = BT,
    long  = behavior tree
}
\DeclareAcronym{CNN}{
    short = CNN,
    long  = convolutional neural network
}
\DeclareAcronym{DVL}{
    short = DVL,
    long  = doppler velocity log
}

\DeclareAcronym{GAN}{
    short = GAN,
    long  = generative adversarial network
}
\DeclareAcronym{GPS}{
    short = GPS,
    long  = global positioning system
}
\DeclareAcronym{HIL}{
    short = HIL,
    long  = hardware-in-the-loop
}
\DeclareAcronym{IMU}{
    short = IMU,
    long  = inertial measurement unit
}
\DeclareAcronym{LBL}{
    short = LBL,
    long  = long baseline
}
\DeclareAcronym{MPC}{
    short = MPC,
    long  = model predictive control
}
\DeclareAcronym{NN}{
    short = NN,
    long  = neural network
}
\DeclareAcronym{NMPC}{
    short = NMPC,
    long  = nonlinear model predictive control
}
\DeclareAcronym{ODE}{
    short = ODE,
    long  = Open Dynamic Engine
}
\DeclareAcronym{OGRE}{
    short = OGRE,
    long  = Object-Oriented Graphics Rendering Engine
}
\DeclareAcronym{OS}{
    short = OS,
    long  = operating system
}
\DeclareAcronym{RCNN}{
    short = RCNN,
    long  = region-based convolutional neural network
}
\DeclareAcronym{RL}{
    short = RL,
    long  = reinforcement learning
}
\DeclareAcronym{ROS}{
    short = ROS,
    long  = robot operating system
}
\DeclareAcronym{ROV}{
    short = ROV,
    long  = remotely operated vehicle
}
\DeclareAcronym{SDF}{
    short = SDF,
    long = simulation description format
}
\DeclareAcronym{SITL}{
    short = SITL,
    long  = software-in-the-loop
}
\DeclareAcronym{SLAM}{
    short = SLAM,
    long  = simultaneous localization and mapping
}
\DeclareAcronym{SOTA}{
    short = SOTA,
    long  = state-of-the-art
}
\DeclareAcronym{UAV}{
    short = UAV,
    long = unmanned aerial vehicle
}
\DeclareAcronym{UE}{
    short = UE,
    long  = Unreal Engine
}
\DeclareAcronym{URDF}{
    short = URDF,
    long  = unified robot description format
}
\DeclareAcronym{URS}{
    short = URS,
    long  = underwater robotic simulator
}
\DeclareAcronym{USBL}{
    short = USBL,
    long  = ultra-short baseline
}
\DeclareAcronym{UUV}{
    short = UUV,
    long  = unmanned underwater vehicle
}
\DeclareAcronym{UVMS}{
    short = UVMS,
    long  = underwater vehicle-manipulator system
}
\DeclareAcronym{VAE}{
    short = VAE,
    long  = variational autoencoder
}
\DeclareAcronym{VSLAM}{
    short = VSLAM,
    long  = visual simultaneous localization and mapping
}
\DeclareAcronym{WHOI}{
    short = WHOI,
    long  = Woods Hole Oceanic Institution
}

\usepackage{chronosys}
\usepackage{etoolbox} 
\usepackage[T1]{fontenc}

\begin{document}
\title{Underwater Robotic Simulators Review for Autonomous System Development}

\author{Sara~Aldhaheri\textsuperscript{1,2},
        Yang~Hu\textsuperscript{1},
        Yongchang~Xie\textsuperscript{1},
        Peng~Wu\textsuperscript{1},
        Dimitrios~Kanoulas\textsuperscript{1},
        and~Yuanchang~Liu\textsuperscript{1}
        \\
    \textsuperscript{1}University College London, London, United Kingdom\\
    \textsuperscript{2}Technology Innovation Institute, Abu Dhabi, United Arab Emirates\\
    \{\textit{sara.aldhaheri.22, yang.hu.24, yongchang.xie.22, wu.peng, d.kanoulas, yuanchang.liu\}}@ucl.ac.uk} 

\maketitle

\begin{abstract}
The increasing complexity of underwater robotic systems has led to a surge in simulation platforms designed to support perception, planning, and control tasks in marine environments. However, selecting the most appropriate \ac{URS} remains a challenge due to wide variations in fidelity, extensibility, and task suitability. This paper presents a comprehensive review and comparative analysis of five state-of-the-art, ROS-compatible, open-source \acp{URS}: Stonefish, DAVE, HoloOcean, MARUS, and UNav-Sim. Each simulator is evaluated across multiple criteria including sensor fidelity, environmental realism, sim-to-real capabilities, and research impact. We evaluate them across architectural design, sensor and physics modeling, task capabilities, and research impact. Additionally, we discuss ongoing challenges in sim-to-real transfer and highlight the need for standardization and benchmarking in the field. Our findings aim to guide practitioners in selecting effective simulation environments and inform future development of more robust and transferable \acp{URS}.

\end{abstract}

\IEEEpeerreviewmaketitle

\section{Introduction}

\begin{figure}[hb!]
    \begin{center}
        \begin{subfigure}[b]{.485\columnwidth}
        \includegraphics[width=\linewidth]{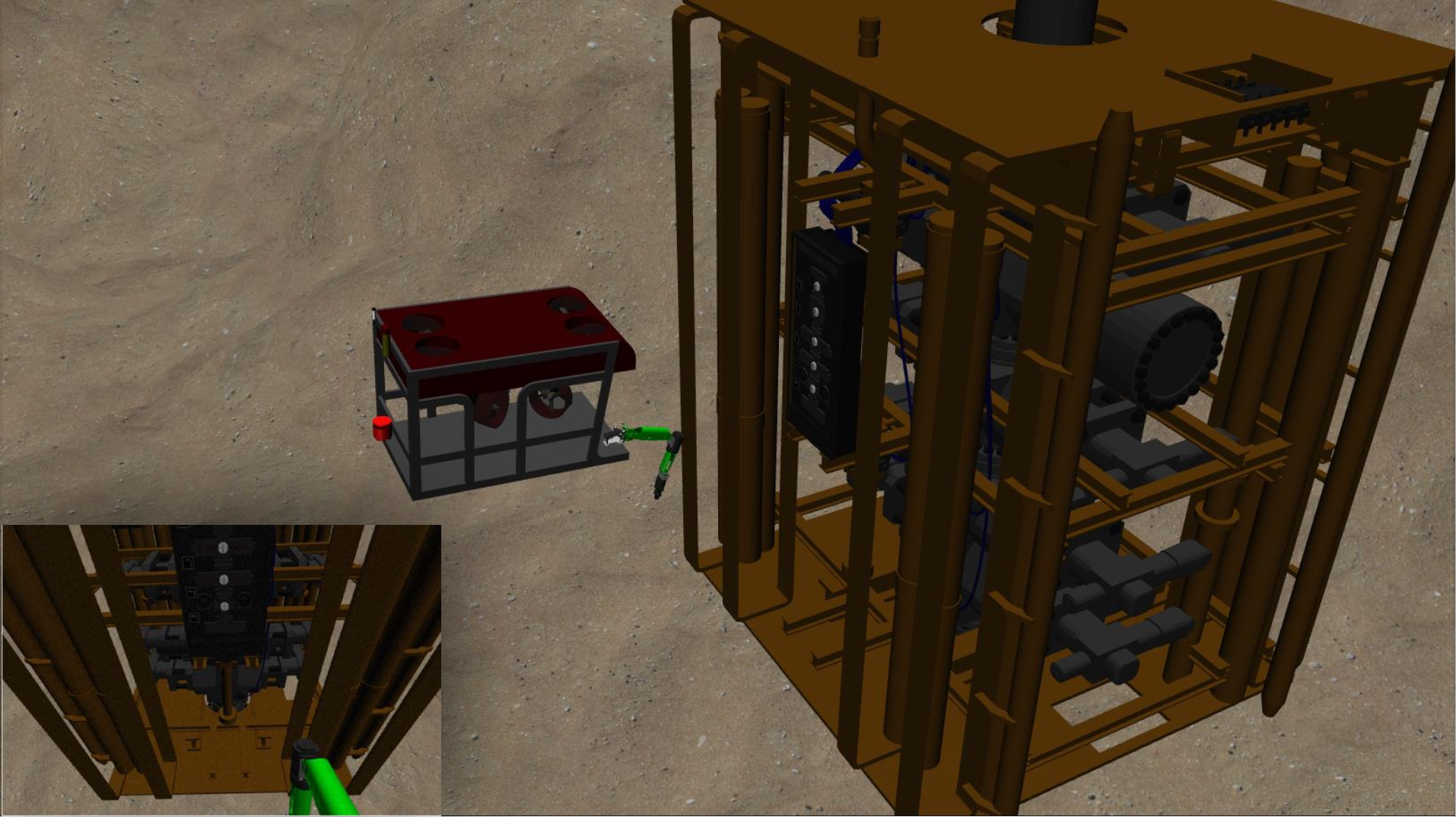}            
        \caption{Project DAVE \cite{zhang2022dave} (2022)}
        \end{subfigure}
        \hfill
        \begin{subfigure}[b]{.485\columnwidth}
        \includegraphics[trim={0 0 0 13.5} , clip, width=\linewidth]{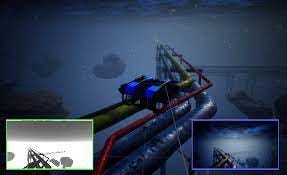}       
        \caption{UNAV-Sim \cite{amer2023unav} (2023)}
        \end{subfigure}
        \hfill
        \hfill
        \begin{subfigure}[b]{.485\columnwidth}
        \includegraphics[width=\linewidth]{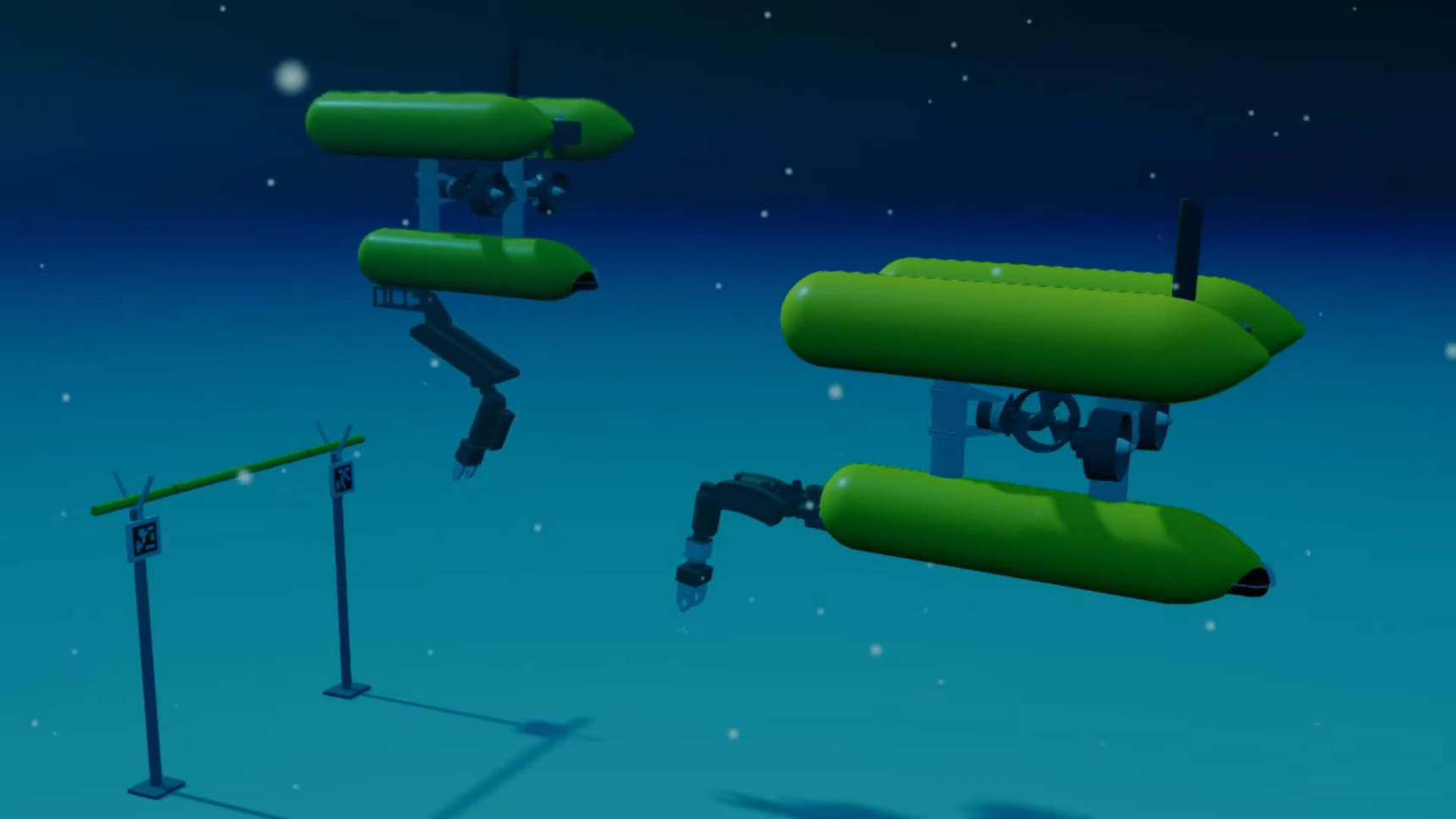}            
        \caption{TWINBOT simulated using Stonefish \cite{cieslak2019stonefish, pi2021twinbot} (2019)}
        \end{subfigure}
        \hfill
        \begin{subfigure}[b]{.485\columnwidth}
        \includegraphics[trim={0 35 0 25} , clip, width=\linewidth]{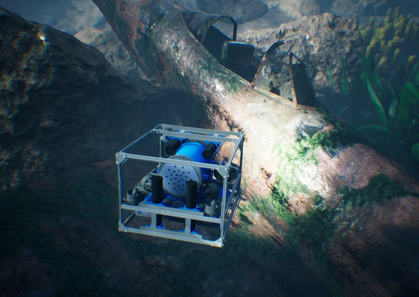}
        \caption{HoloOcean \cite{potokar2022holoocean} (2022)}
        \hfill
        
        \end{subfigure}
    \end{center}
    \caption{State-of-the-art underwater robotic simulators.}
    \label{fig:simulators}
\end{figure}

Underwater technology advancements have driven significant growth in autonomous marine systems over the past decade \cite{neira2021review, christensen2022recent, wang2022development}. The expansion in this field has led to the release of numerous simulator systems (see \Cref{fig:simulators}), which enhances virtual realism while offering greater adaptability offered by simulation. Despite the surge in \acp{URS}, there has not been a comprehensive review that critically surveys their capabilities in certain sub-sea applications, contributing to the high entry barrier for research in this field. Existing studies are either outdated, narrowly scoped, or do not account for the increasing complexity of modern underwater robotics \cite{matsebe2008review,cook2014survey,collins2021review,ciuccoli2024underwater}. As these toolkits grow, there is a pressing need for a study that not only addresses the full range of emerging features but also evaluates their real-world impact on underwater robotic applications, ensuring the selection the most effective simulation platforms for their use case.

Therefore, we explore the literature to identify the key factors required for accurately modeling marine scenarios and evaluate the applicability of state-of-the-art \acp{URS}\footnote{An online library of the surveyed literature is available at \url{https://frl-ucl.github.io/projects/underwater-robotic-simulators.html}}, with a focus on particular aspects such as capabilities, research impact, and sim-to-real transition. The goal of this work is to move beyond a general comparative review, offering an in-depth discussion on the strengths, limitations, and ideal use case of each \ac{URS}. In doing so, we provide actionable guidance to practitioners in selecting the most suitable tools for their needs. 

Our analysis begins with an overview of key features and architectural components for establishing effective \acp{URS} in \Cref{sec:overview}. Then, we examine leading simulators, emphasizing their functionalities that position them as promising candidates for future underwater robotics research and development in \Cref{sec:comparativereview}. Next, we highlight recent endeavors that tackles sim-to-real transfers and the ongoing challenges faced when transitioning from simulation to field deployment in \Cref{sec:sim2real}. Finally, we present a discussion of current limitations and potential directions for improvement in \Cref{sec:discussion}, followed by concluding insights in \Cref{sec:conclusion}.


\section{Underwater Robotic Simulation Overview}
\label{sec:overview}
\Acp{URS} are software platforms that emulate the behavior, dynamics, and interactions of underwater robotic systems within a virtual environment. They typically incorporate elements such as realistic physics engines, interactive control interfaces, detailed environment modeling, and user-defined mechanisms. These platforms offer extensive experimental capabilities in an field of research that would normally involve expensive equipments and difficult-to-access locations. The complexities of the underwater environment, characterized by dynamic currents, varying pressure levels, low visibility, and challenging communication conditions, further increases required resources and risk of physical testing. Thus, modeling underwater environment is an essential step towards the development of marine systems to provide a controlled environment where critical algorithms can be tested, refined, and validated before real-world deployment. However, unlike the progress seen in modeling non-marine scenarios (e.g., autonomous driving \cite{dosovitskiy2017carla}, aerial robotics  \cite{shah2018airsim}, locomotion \cite{todorov2012mujoco}, etc.), the development of use-case-specific subsea simulators has only recently been catching up, largely due to the unique challenges associated with accurately replicating and validating the underwater environment. These challenges include simulating fluid dynamics, sensor noise, and the interactions between vehicles and complex, unstructured underwater terrains. While advancements have been made, many existing simulators lack the fidelity and support required for the diverse range of underwater applications, highlighting the need for more robust and specialized platforms to bridge this gap. 

Therefore, this section offers an overview of the key components commonly found in \acp{URS}, providing insight into the underlying mechanisms of these simulators in addition to clarifying the selection criteria adopted in this manuscript.

\subsection{Core modules of state-of-the-art \acp{URS}}
\begin{figure}
    \centering
    \includegraphics[width=\linewidth]{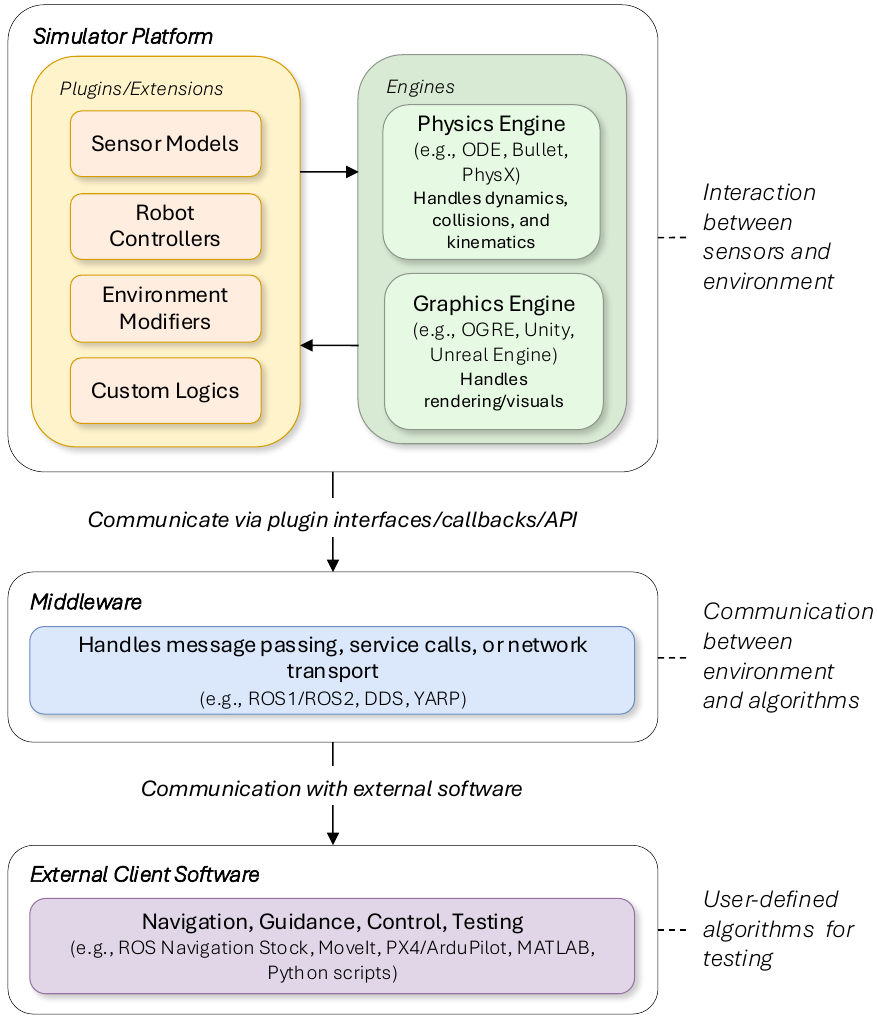}
    \caption{Foundational framework for robotic simulators.\vspace{-1em}}
    \label{fig:framework}
\end{figure}

To model underwater environments, simulators must account for dynamic, variable properties using architectures that enable robust interaction between the environment and internal systems. \Cref{fig:framework} illustrates a foundational framework for robotic simulators, highlighting the primary modules and their relationships which are further discussed in this section.

\subsubsection{Understanding the simulator framework}

\paragraph{Plugins / Extensions} These are modular components integrated with the simulator to provide specialized functionalities of real system counterparts (e.g., sensors, motors, hydrodynamics, etc.). Sensor plugins in \acp{URS} can generally be grouped into two main categories: \emph{optical} and \emph{acoustic} devices. Common optical sensors include standard \emph{RGB cameras} and \emph{LIDARs}, often used for perception, mapping, and navigation tasks. On the acoustic side, widely supported devices include \emph{multibeam sonars}, \emph{forward-looking sonars (FLS)}, \emph{side-scan sonars (SSS)}, \emph{micro-modem-based acoustic positioning systems} like \emph{\ac{USBL}} and \emph{\ac{LBL}}, and \emph{\acp{DVL}}, which estimate velocity relative to the seabed. These plugins simulate real-world sensor behavior and are essential for replicating underwater perception and localization in GPS-denied environments. Other frequently used sensor types include \emph{\acp{IMU}}, \emph{magnetometers}, and \emph{pressure sensors}.

On the actuator side, plugins are implemented for \emph{thrusters} that generate propulsion forces in multiple directions to maneuver, hover, or maintain position, and for control surfaces or adjustable fins that create hydrodynamic forces to enhance stability and maneuverability, especially in vehicles equipped with control \emph{fins} or \emph{rudders}. With regards to \acp{UVMS}, this could include plugins for \emph{servo} or \emph{DC motors}, considering the three types of joints commonly found in such systems: prismatic, revolute, and fixed.

These sensor and actuator plugins often incorporate configurable parameters (e.g., noise levels, response delays, hydrodynamic coefficients) to better approximate real‐world performance. This is one area where researchers are able to fill the sim-to-real gap by adding disturbance that real sensors would likely face  \cite{varshosaz2024monitoring}. Environment modifier plugins enable customization of a simulation’s physical surroundings. They interface with the physics engine by \emph{reading} states like position, velocity, and collisions, and \emph{applying} forces or torques based on sensor feedback and actuator commands. This bidirectional flow ensures realistic simulations that accurately reflect the dynamics and control of unmanned underwater vehicles.

\paragraph{Physics Engine} In the domain of robotic simulations, various physics engines have been used to emulate real‐world dynamics and kinematics~\cite{collins2021review}. \emph{Bullet}, \emph{PhysX}, \emph{\ac{ODE}}, and \emph{Unity Physics} are among the most widely adopted engines in both academia and industry. They provide essential features such as gravity, friction, rigid‐body dynamics, collision detection, and — in some cases — extensions for fluid and underwater modeling. By leveraging these engines, developers can build immersive and realistic simulation scenarios that capture phenomena like the Coriolis effect, wave action, or complex hydrodynamic interactions. In turn, sensor and actuator plugins feed their data into the physics engine, ensuring the simulation reflects the actual physics of underwater robotics. 

Within a \ac{URS}, each simulated object possesses physical attributes such as mass, friction, density, and buoyancy coefficients to mimic real‐life underwater interactions. These properties are typically described in standardized file formats known as the \ac{URDF} for the robot systems and \ac{SDF} for the world models. As a high‐level toolset, the physics engine updates the simulator state at each time step, calculating how objects move and react to applied forces or environmental factors.

\paragraph{Visual Rendering or Graphics Engine} While the physics engine governs how objects \emph{move} and \emph{interact}, the graphics engine renders these interactions in a visually coherent and immersive \ac{3D} environment. Commonly used rendering engines include \emph{\ac{OGRE}}, \emph{Ogre3D}, \emph{Unity}, \emph{\ac{UE}}, and \emph{Blender}. They offer robust frameworks that leverage advanced features such as ray tracing, shader programming, and physics‐based rendering to create realistic scenes \cite{choi2021physics}. 

Visual realism is particularly important for tasks like computer vision, object detection, and operator training in marine robotic systems. Rendering engines thus support environmental effects such as: light scattering and refraction, turbidity and caustics, and lastly, underwater textures. With the option to tune these scenarios, practitioners can emulate specific mission conditions, such as murky waters for sediment-heavy environments, shifting light patterns for varying depths, or complex seabed structures for navigation and manipulation as seen in \cite{gaspar2023feature,ruscio2023autonomous,gaspar2023visual}.

Finally, simulation platforms seamlessly integrate graphics and physics engines with the plugin extensions to create dynamic environments. These components interface with \emph{middleware} for real-time rendering and data exchange, while also translating inputs from \emph{external client modules} where users can define and test their own algorithms. By blending high-fidelity visuals with accurate physics and sensor/actuator models, simulators offer an immersive environment to validate custom user algorithms before deployment.



\subsubsection{Modeling the System}
The dynamics of underwater robots can be primarily categorized into rigid-body dynamics, governed by Newton-Euler physics, and hydrodynamic and hydrostatic effects, arising from the surrounding fluid. The simulator accounts for hydrodynamics through added mass and damping forces. Hydrodynamic added mass acts as a virtual mass added to the system when a body accelerates or decelerates, displacing a portion of the nearby fluid, whereas hydrodynamic damping primarily includes potential damping, skin friction, wave drift damping, and vortex shedding effects. This dynamic model is expressed in the following equation \cite{fossen1999guidance}:

\begin{equation}
\label{eq:auv_dynamics_tworows_underbrace}
\begin{aligned}
\boldsymbol{\tau} &= \underbrace{\mathbf{M_{RB}} \dot{\mathbf{v}} + \mathbf{C_{RB}}(\mathbf{v}) \mathbf{v}}_{\text{Rigid-Body Terms}} \\
&\quad + \underbrace{\mathbf{M_A} \dot{\mathbf{v}} + \mathbf{C_A}(\mathbf{v}) \mathbf{v} + \mathbf{D}(\mathbf{v}) \mathbf{v} + \mathbf{g}(\boldsymbol{\eta})}_{\text{Hydrodynamic and Hydrostatic Terms}}
\end{aligned}
\end{equation}

where \(\boldsymbol{\tau}\) denotes the force/moment vector; \(\mathbf{v}\) represents the velocity; \(\mathbf{M_{RB}}\) and \(\mathbf{C_{RB}}(\mathbf{v})\) are the rigid-body mass and Coriolis/centripetal matrices; \(\mathbf{M_A}\) and \(\mathbf{C_A}(\mathbf{v})\) are the added mass and hydrodynamic Coriolis/centripetal matrices; \(\mathbf{D}(\mathbf{v})\) is the damping matrix for fluid resistance; and \(\mathbf{g}(\boldsymbol{\eta})\) is the gravitational/buoyant force vector.

This dynamics model accurately captures the complex dynamics of underwater robots, integrating rigid-body motion with hydrodynamic and hydrostatic effects. It enables realistic simulation of how forces, water resistance, and buoyancy influence movement, essential for testing control systems and predicting real-world behavior efficiently.

\subsubsection{Adding disturbances / noises (e.g., currents, waves, etc.)}
\label{sec:noise}

Modeling disturbances is a crucial capability, particularly when ground truth data is difficult or impossible to obtain. Disturbances can arise from various sources, including environmental factors and sensor noise. 

Environmental disturbances include factors such as waves, currents, and water clarity, which can affect the motion, navigation, and perception capabilities of underwater vehicles. A common approach to simulating these disturbances is the introduction of Gaussian noise, which helps replicate the unpredictability of natural underwater environments \cite{deng2024simulating, hu2024disturbance}. More specifically, an error model based on the first-order Gauss–Markov process is employed to capture temporal correlations and inherent variability, providing a more realistic representation of underwater dynamics:
\begin{subequations}
\begin{align}
    \mathcal{X} &= S + n + G_s \label{eq:1a} \\[6pt]
    \dot{n} &= -\frac{1}{\tau} + G_b \label{eq:1b}
\end{align}
\end{subequations}
where $\mathcal{X}$ represents the system state, $S$ is the true signal (e.g., the robot's intended position), $n$ is the noise term representing the disturbance, $\tau$ is the time constant determining the correlation decay, and $G_b$ is Gaussian white noise. This model is particularly useful for simulating dynamic environmental factors like ocean currents or wave-induced motions, as it accounts for the temporal correlation of disturbances over time. 

Sensor noise is also essential for realistic underwater perception simulation, especially under low-visibility conditions where sonar dominates. As described in \cite{song2025oceansim,potokar2022holoocean,akkaynak2018revised}, imaging sonar noise typically includes both additive and multiplicative components to model background interference and signal-dependent distortions, respectively. Beyond these, other sensor disturbances can further impact realism and measurement fidelity. These can include quantization noise from limited resolution, temporal drift due to thermal or electronic instability, and motion-induced distortions arising from latency or non-instantaneous acquisition. Environmental effects such as backscatter, multipath interference, and medium-dependent attenuation (e.g., turbidity or salinity gradients) can introduce complex artifacts or signal degradation. Spatial inconsistencies may also result from beam pattern non-uniformities or occlusions in complex geometries. Additionally, interference from external acoustic sources and calibration errors, such as gain mismatch or orientation drift, can compound measurement uncertainty. Incorporating these effects is critical for developing robust perception algorithms that generalize to real-world deployments, effectively narrowing the gap between simulation and reality.

\subsection{Key Criteria for Simulator Selection}
Three critical factors guided the authors' decision when choosing the simulators explored: \ac{ROS} support, maintenance, and licensing.

\subsubsection{\textbf{ROS Support}} Integrated \ac{ROS} support is essential because it provides a flexible, modular framework for simulating sensors, actuators, and control systems with high fidelity. By leveraging \Ac{ROS}, developers can simulate marine operations with high fidelity, including navigation, manipulation, and environmental interaction, while benefiting from its extensive library of prebuilt packages and robust community support. \Ac{ROS} enables seamless communication between software components, facilitating modular development and rapid prototyping of complex systems. Moreover, \Ac{ROS}'s compatibility with various programming languages (e.g., Python, C++) and middleware (e.g., DDS in ROS2) supports diverse research needs, including real-time systems, distributed robotics, autonomous navigation, machine learning integration, and advanced control algorithm development. Therefore, this review focuses on simulators with native \ac{ROS} support because they enable precise and flexible simulation of complex underwater operations.

\subsubsection{\textbf{Maintenance}} Another critical factor in selecting the \acp{URS} reviewed in this manuscript is the availability of their maintenance support, particularly through active user forums and community engagement. Priority was given to simulators that demonstrated active upkeep, updates to dependencies, and responsive developer community during the simulator's adoption period. Additionally, a conservative approach was taken with newly released simulators, as their long-term support and sustainability remain uncertain. Furthermore, we assess each simulator’s lifecycle and reliability by examining both its historical stability and current maintenance activity, and we consider the feasibility of migrating to ROS2 or adopting Docker for improved future compatibility and simpler deployment.


\subsubsection{\textbf{Open-source vs Proprietary}} The choice between proprietary and open-source underwater robotic simulators hinges on the specific needs of researchers and industry professionals. Open-source platforms, such as those further detailed in \Cref{sec:comparativereview}, thrive on accessibility and customizability, enabling users to extend their capabilities for a wide range of applications. This makes them particularly attractive for academic research and exploratory projects \cite{abdullah2024human}. In contrast, proprietary simulators offer targeted functionality and professional support, making them invaluable for specialized tasks. 

While proprietary tools often come with polished features and developer support, open-source solutions are more cost effective, accessible and adaptable, which foster collaboration and innovation across diverse developers and their use cases. In this regard, this review focuses on open-source simulators released in recent years.


\section{Comparative Review of State-of-the-art Underwater Robotic Simulation}
\label{sec:comparativereview}

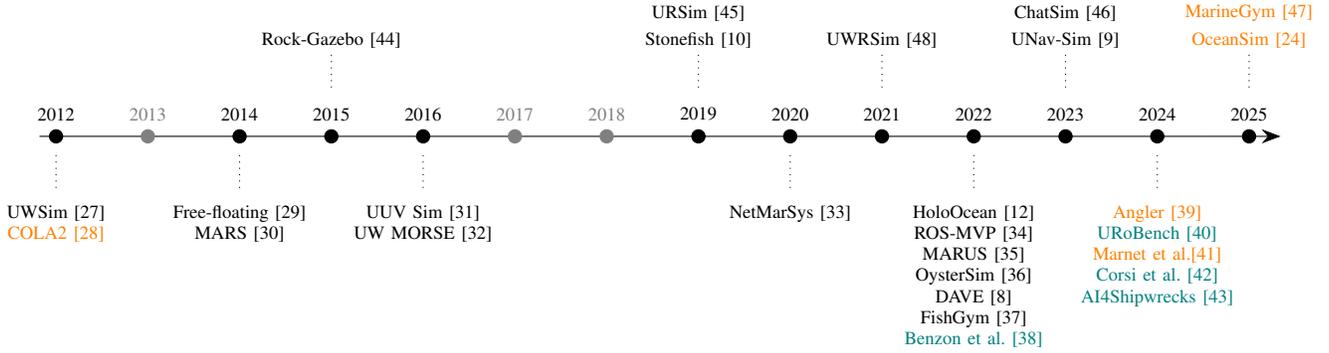
\begin{figure*}[h!]
\centering
\begin{tikzpicture}[node distance=1.5cm]
    \draw[thin] (0.5,0) -- (16,0); 
    \draw[thin, -{Stealth[scale=1.5]}] (16,0) -- (17,0); 

    \foreach \year/\event [count=\i] in {
        2012/{UWSim \cite{dhurandher2008uwsim}\\ \textcolor{orange}{COLA2 \cite{palomeras2012cola2}}},
        \textcolor{gray}{2013}/, 
        2014/{Free-floating \cite{conti2017free}\\ MARS \cite{tosik2014mars}},
        {2015}/{}, 
        2016/{UUV Sim \cite{manhaes2016uuv}\\UW MORSE \cite{henriksen2016uw}},
        \textcolor{gray}{2017}/,
        \textcolor{gray}{2018}/,
        2019/{},
        2020/{NetMarSys \cite{garg2020netmarsys}},
        2021/,
        2022/{HoloOcean \cite{potokar2022holoocean}\\ROS-MVP \cite{gezer2022working} \\MARUS \cite{lonvcar2022marus}\\OysterSim \cite{lin2022oystersim}\\DAVE \cite{zhang2022dave}\\FishGym \cite{liu2022fishgym}\\ \textcolor{teal}{Benzon et al. \cite{von2022open}}},
        2023/,
        2024/{\textcolor{orange}{Angler \cite{palmer2024angler}}\\\textcolor{teal}{URoBench \cite{huang2024urobench}}\\\textcolor{orange}{Marnet et al.\cite{marnet2024bridging}}\\\textcolor{teal}{Corsi et al. \cite{corsi2024aquatic}}\\\textcolor{teal}{AI4Shipwrecks \cite{sethuraman2024machine}}},
        2025/
    } {
        \node[above, yshift=2.5pt, font=\scriptsize] at (\i*1.22-0.5,0) {\year}; 
        \ifodd\i
             \ifnum\i=7
                \filldraw[gray] (\i*1.22-0.5,0) circle (2.5pt); 
                
            \else
                \draw[dotted] (\i*1.22-0.5,-0.25) -- (\i*1.22-0.5,-0.7); 
                \filldraw (\i*1.22-0.5,0) circle (2.5pt); 
            \fi
            
        \else
            \ifboolexpr{test {\ifnumcomp{\i}{=}{2}} or test {\ifnumcomp{\i}{=}{6}}}
            {
                \filldraw[gray] (\i*1.22-0.5,0) circle (2.5pt); 
            }
            {
                \filldraw (\i*1.22-0.5,0) circle (2.5pt); 
            }
        \fi        
        \node[below, align=center, anchor=north, yshift=-22.5pt, font=\scriptsize] at (\i*1.22-0.5,0) {\event}; 
    }

    \node[above, yshift=30pt, font=\scriptsize] at (4*1.22-0.5,0) {Rock-Gazebo \cite{watanabe2015rock}}; 
    \draw[dotted] (4*1.22-0.5,0.65) -- (4*1.22-0.5,1.1); 
    \node[above, yshift=30pt, font=\scriptsize] at (8*1.22-0.5,0) {Stonefish \cite{cieslak2019stonefish}}; 
    \node[above, yshift=40pt, font=\scriptsize] at (8*1.22-0.5,0) {URSim \cite{katara2019open}}; 
    \draw[dotted] (8*1.22-0.5,0.65) -- (8*1.22-0.5,1.1); 
    \node[above, yshift=40pt, font=\scriptsize] at (12*1.22-0.5,0) {ChatSim \cite{palnitkar2023chatsim}}; 
    \node[above, yshift=30pt, font=\scriptsize] at (12*1.22-0.5,0) {UNav-Sim \cite{amer2023unav}}; 
    \draw[dotted] (12*1.22-0.5,0.65) -- (12*1.22-0.5,1.1); 
    \node[above, yshift=30pt, font=\scriptsize] at (14*1.22-0.5,0) {\textcolor{orange}{OceanSim \cite{song2025oceansim}}}; 
    \node[above, yshift=40pt, font=\scriptsize] at (14*1.22-0.5,0) {\textcolor{orange}{MarineGym \cite{chu2024marinegym}}}; 
    \draw[dotted] (14*1.22-0.5,0.65) -- (14*1.22-0.5,1.1); 
    \node[above, yshift=30pt, font=\scriptsize] at (10*1.22-0.5,0) {UWRSim \cite{chaudhary2021development}}; 
    \draw[dotted] (10*1.22-0.5,0.65) -- (10*1.22-0.5,1.1); 

\end{tikzpicture}
\caption{\textcolor{black}{A timeline of \ac{URS} (black), sim-to-real transfer frameworks (\textcolor{orange}{orange}), and benchmarking tools (\textcolor{teal}{teal}) released in recent years.}}
\label{fig:timeline}
\end{figure*}
\begin{table*}[h!]
    \centering
    \renewcommand{\arraystretch}{1.2} 
    \renewcommand\tabularxcolumn[1]{m{#1}}    
    \caption{Comparative analysis based on key technological features where L stands for Linux, W is for Windows, and O is for macOS.}
    \label{tab:comparison2a}
    \begin{tabularx}{\textwidth}{>{\centering\arraybackslash}X>{\centering\arraybackslash}X>{\centering\arraybackslash}X>{\centering\arraybackslash}X>{\centering\arraybackslash}X>{\centering\arraybackslash}X>{\centering\arraybackslash}X} 
        \toprule
        \textbf{Simulator} & \textbf{Release Year $\uparrow$} & \textbf{OS (L/W/O)} & \textbf{ROS Integration} & \textbf{Physics Engine} & \textbf{Graphics Engine} \\
        \midrule
        
        Stonefish \cite{cieslak2019stonefish} & 2019 & LWO & C++ library with ROS interface & Bullet & OpenGL \\
        DAVE      \cite{zhang2022dave} & 2022 &  L  & ROS1 & Gazebo Physics Library &  OpenGL \\
                    
        HoloOcean \cite{potokar2022holoocean}& 2022 & LW  & ROS2 Wrapper & PhysX & \ac{UE}4.27, \ac{UE}5.3(Pre-release)  \\

        MARUS  \cite{lonvcar2022marus} & 2022 & LW  & ROS1,2 & Unity3D & Unity3D \\

        UNav-Sim  \cite{amer2023unav} & 2023 & LW  & ROS1,2 & Fast Physics Engine by AirSim & \ac{UE}5.1 \\

        \bottomrule
    \end{tabularx}
    
\end{table*}


\begin{table*}[h!]
    \centering
    \renewcommand{\arraystretch}{1.2} 
    \renewcommand\tabularxcolumn[1]{m{#1}}    
    \caption{Comparative analysis based on task capabilities. A dash represents the lack of explicit documentation.}
    \label{tab:comparison2b}
    \begin{tabularx}{\textwidth}{>{\centering\arraybackslash}X>{\centering\arraybackslash}X>{\centering\arraybackslash}X>{\centering\arraybackslash}X>{\centering\arraybackslash}X>{\centering\arraybackslash}X>{\centering\arraybackslash}X} 
        \toprule
        \textbf{Simulator} & \textbf{Manipulator Capabilities} & \textbf{Multi-vehicle Support} & \textbf{Acoustic Communications} & \textbf{Perception Task Realism}\\
        \midrule
        
        Stonefish \cite{cieslak2019stonefish} & 
        \checkmark & \checkmark & \checkmark & High fidelity \\

        DAVE  \cite{zhang2022dave}    &\checkmark & \checkmark & \xmark & Moderate fidelity  \\

        HoloOcean \cite{potokar2022holoocean} & \xmark & \checkmark &\checkmark & High fidelity \\
                    
        MARUS  \cite{lonvcar2022marus} & \xmark  & \xmark & \checkmark & Moderate fidelity \\

        UNav-Sim  \cite{amer2023unav} & \xmark  & \xmark & \xmark & High fidelity\\
        \bottomrule
    \end{tabularx}
\end{table*}

Although numerous state-of-the-art \acp{URS} have been developed (see \Cref{fig:timeline}), this comparative study focuses on five of the more recent and cited simulators that exhibit extensibility and potential for long-term sustainability. These simulators are also distinct in terms of their underlying engines and capabilities, albeit they share similarities in the types of sensors they emulate. Detailed comparisons are provided in \Cref{tab:comparison2a} and \Cref{tab:comparison2b}, where simulators are evaluated across several key dimensions. These include their year of release, supported \acp{OS}, and integration with \ac{ROS}, which are crucial for assessing compatibility and relevance. Advanced features such as manipulator support, multi-vehicle simulation, acoustic communications, and realism are considered for their impact on simulation accuracy and versatility. The rest of this section explores the simulators in in greater depths of their research application.



\subsection{Stonefish}
Stonefish is an extensible underwater robotic simulator that has been widely utilized in research and various applications. Its high-fidelity, \ac{ROS}-compatible framework offers many opportunities to advance underwater robotics research by accurately simulating \ac{AUV} dynamics and sensor-rich platform. Stonefish defines a simulation using \textit{scenarios} which are XML files detailing the environment and the robot.  It has showcased its effectiveness across perception, planning and control research. For example, in adaptive visual exploration, Stonefish enabled researchers to generate realistic seafloor imagery, complete with textured representations of seagrass meadows and variable terrains, which allowed for precise quantitative evaluation of mapping algorithms and decision-time adaptive replanning strategies \cite{guerrero2021adaptive}. 
Similarly, for bathymetric surveying using imaging sonar and neural volume rendering, Stonefish played a critical role by synthesizing realistic forward-looking sonar data along survey trajectories, thereby enabling the development and validation of self-supervised reconstruction methods that bridged the gap between low-resolution prior maps and high-resolution field data \cite{xie2024bathymetric}. For the development of the low-cost, portable ALPHA \ac{AUV}, the framework provided a platform to simulate vehicle dynamics and sensor interactions, accelerating control system development and ensuring that guidance, navigation, and control algorithms were thoroughly vetted before field deployment \cite{zhou2022acrobatic}. 
In the area of \ac{RL} for hydrobatic maneuvering, Stonefish offered a safe and computationally efficient simulation space where \ac{RL} agents could be trained to handle the non-linear, coupled dynamics of agile \acp{AUV}, yielding performance comparable to classical PID controllers and exposing sim-to-real transfer challenges that guided further refinements \cite{wozniak2024using}. Furthermore, in the study on learning the ego-motion of underwater imaging sonar through \ac{CNN} and \ac{RCNN} architectures, the framework was essential for generating vast amounts of labeled synthetic sonar imagery under varied simulated conditions, which bolstered the training process and enhanced the models' ability to generalize to real-world scenarios \cite{munoz2024learning}. 

Finally, Stonefish has been extended to further support marine machine learning research through the addition of new sensors like \emph{event-based camera}, \emph{thermal camera}, and \emph{optical flow sensor} \cite{grimaldi2025stonefish} resulting in the most extensive sensor suite to the best of the authors knowledge. It also now features \emph{visual light communication} and an automated annotation tool allowing for efficient dataset labeling for underwater robotics perception tasks.


\subsection{DAVE Project}
Building on the UUV Simulator and \ac{WHOI} Deep Submersible Laboratory's Gazebo package \verb|ds_sim| \cite{manhaes2016uuv,vaughnDSSim}, the DAVE Aquatic Virtual Environment Project offers a platform with fundamental sensors for testing underwater robotic solutions \cite{zhang2022dave}. 
In multiple studies, the DAVE simulation has proven integral offering an edge with works towards underwater manipulation (e.g. bimanual manipulation). For instance, in underwater calibration arm development for aquaculture, DAVE allowed precise simulation of the kinematic control and environment-specific physics affecting manipulator arms, crucial for automating camera calibration processes that significantly enhance aquaculture management \cite{voskakis2024modeling}. Similarly, DAVE facilitated scalable underwater assembly operations by enabling simulations that guided the positioning and manipulation tasks of \acp{AUV} using reconfigurable visual fiducial markers, improving both task efficiency and localization accuracy within complex underwater assembly scenarios \cite{lensgraf2024scalable}. 
Furthermore, DAVE's robust physics and sensor modeling capabilities supported advanced underwater SLAM solutions, specifically for integrating imaging sonar data with visual-inertial sensors, improving robustness against visual degradation during localization tasks \cite{pan2025russorobustunderwaterslam}. Additionally, predictive modeling of underwater vehicle manipulator dynamics was effectively achieved through time-series methods tested within DAVE’s realistic simulation environment, enhancing the capability to forecast \ac{UVMS} behaviors under harsh environmental conditions typically found near aquaculture sites \cite{chan2023predicting}. 
Finally, DAVE contributed to developing efficient reactive obstacle avoidance algorithms utilizing forward-looking sonar for real-time \ac{3D} navigation, enabling \acp{AUV} to maneuver safely through cluttered underwater environments \cite{mane2024eroas3defficientreactive}. Collectively, these diverse studies showcase DAVE's central role in enabling robust, precise, and scalable solutions across various domains of underwater robotics research.  

\subsection{HoloOcean}
HoloOcean offers an advantage for projects focusing on realism with the ability to add new environments or assets via \ac{UE}’s editor \cite{potokar2022holoocean,potokar2024holoocean}. Also, its Python API makes it straightforward to script scenarios and customize vehicle dynamics. Unlike ROS-based simulators, HoloOcean’s installation is simple (\verb|pip install|), and it avoids external dependencies, lowering the barrier to use. It is built on the Holodeck framework and supports multiple agents and its high-fidelity environment has influenced several research studies especially in the field of perception. For instance in OceanChat, HoloOcean serves as the foundation for HoloEco which generates photorealistic underwater scenes for testing \ac{AUV} control via natural language commands. This leads to improved success rates and faster computation \cite{yang2023oceanchat}. In another study on neural implicit surface reconstruction, researchers use imaging sonar data from HoloOcean. The high-fidelity sonar simulation helps reconstruct detailed \ac{3D} surfaces. This work overcomes the elevation ambiguity common in sonar measurements \cite{qadri2023neural}. In Actag, HoloOcean enables the simulation of opti-acoustic fiducial markers, demonstrating precise localization and mapping through the fusion of visual and acoustic data \cite{norman2023actag}. HoloOcean also benefits research in navigation and filtering. In a work on invariant filtering for underwater inertial navigation/\ac{DVL}/position system navigation, its realistic sensor models and dynamic fidelity support accurate state estimation \cite{wang2024invariant}. 

\subsection{MARUS}
\underline{MA}rine \underline{R}obotic \underline{U}nity \underline{S}imulator (MARUS) is a Unity3D-based tool that extends the physics engine by including its own water body physics. Using a game engine as a robotic modeling tool allows for GPU utilization offering parallelization of complex computations and high-fidelity visual rendering, while its key feature, the gRPC-enabled bidirectional communication with ROS and other backends, facilitates realistic sensor simulation and real-time algorithm testing. Additionally, MARUS offers automated tools for camera and point-cloud annotation, reducing efforts of labeled dataset generation for machine learning.

Collaborative Aquatic Position System's proof-of-principle experiments were carried out in MARUS's Unity-based nuclear fuel pool environment, wherein the \ac{URS} supplied realistic multi-beam sonar returns, and depth readings, while hydrodynamic forces were modeled by the Dynamic Water Physics asset. Through the discussed gRPC bridge to ROS, MARUS enabled synchronous sharing of ground-truth poses for its vehicles and sensor data, facilitating quantification of localization accuracy with below 0.2 m root mean square error trajectories under controlled but realistic conditions \cite{lennox2024collaborative}.

For the ROADMAP diver-robot navigation project, MARUS's sonar simulator was leveraged to generate large volumes of synthetic forward-looking sonar imagery from multiple viewpoints in a cluttered, mapped environment. This functionality allowed the pre-training and validation of diver-detection and map-relative localization algorithms without the prohibitive logistical overhead of underwater field trials. MARUS thus provides the simulation infrastructure required to integrate precise \ac{AUV} pose estimates with sonar observations to deliver centimeter‑level diver localization for real‑time heads‑up display augmentation \cite{nadj2022towards}.

These applications highlight MARUS's capability for high-fidelity and accurate perception modeling, particularly in complex and data-intensive underwater scenarios.

\subsection{UNav-Sim}

One of the newest \ac{URS}, UNav-Sim is a high-fidelity environment that integrates with both ROS1 and ROS2. It features realistic physics, advanced rendering, and an autonomous vision-based navigation stack, enabling robots to navigate using visual data while addressing challenges in \ac{VSLAM}. To further support research in visual localization, UNav-Sim introduces a benchmarking framework, demonstrated through a pipeline tracking scenario that evaluates state-of-the-art methods such as ORB-SLAM3 \cite{campos2021orb} and TartanVO \cite{wang2021tartanvo}. UNav-Sim's control paradigm investigated through Blue Robotics BlueROV Heavy vehicle. Recently, this simulator has been used for robotic planning works that investigate the use of \acp{BT} \cite{aubard2024mission}. UNav-Sim allowed the testing of an autonomous pipeline inspection framework by evaluating \ac{BT}-based mission planning in a realistic underwater environment, ensuring robots can react to unexpected events. The simulator also validated an AI-driven pipeline inspection algorithm, using simulated visual and sonar data to assess its performance. Additionally, safety verification was demonstrated by running BehaVerify-checked mission plans in UNav-Sim, confirming their reliability in dynamic conditions.

\begin{figure}
    \centering
    \includegraphics[width=\linewidth]{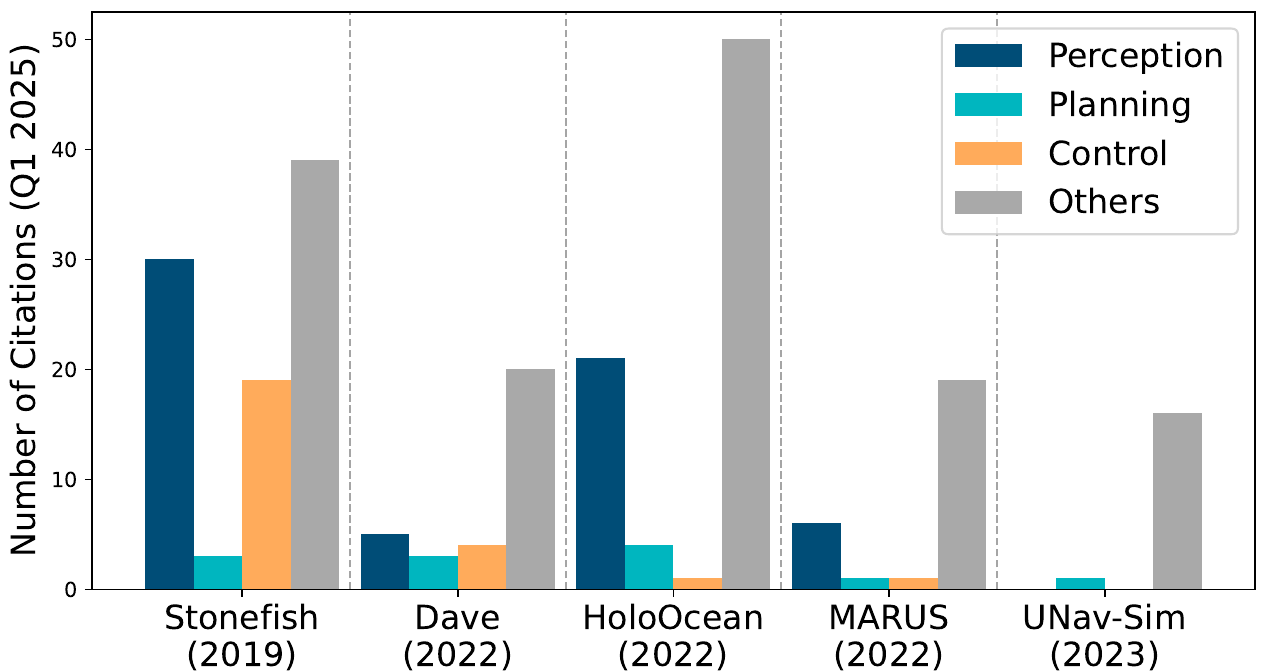}
    \caption{Research area distribution of publications that utilize the five reviewed simulator with the simulators' release years in parenthesis.\vspace{-1em}}
    \label{fig:breakdown}
\end{figure}

To further explore the practical applications of these simulators, we examined over 240 existing studies that have utilized them as testing platforms. Specifically, we analyzed their use across three primary domains: perception, planning, and control. Additionally, we identified a separate category, labeled \emph{Others}, which includes studies that mention these simulators but do not necessarily employ them for testing. \Cref{fig:breakdown} illustrates the total number of mentions for each simulator, with newer simulators appearing toward the rightmost side of the graph.


\section{Bridging Sim-to-Real Gaps in Underwater Robotics}
\label{sec:sim2real}
A core challenge in marine projects is sample efficiency, the requirement for a large volume of training data to effectively train models or agents. Gathering such comprehensive datasets is both costly and time-consuming. As a result, researchers are increasingly relying on sim-to-real approaches, which can involve generating synthetic data \cite{alvarez2023mimir}, training control policies in simulation with software-in-the-loop \cite{palmer2024angler}, employing physics-based simulations to accurately model underwater dynamics and sensor characteristics \cite{song2025oceansim}, etc. However, the transfer of these models, policies, and agents still encounters translational issues that delineates the persistent sim-to-real gap. Advances in perception, control, and planning have significantly improved transferability, yet challenges remain. In perception, high-fidelity sensor simulation and synthetic datasets are key. For example, MIMIR-UW delivers multi-sensor synthetic data with realistic effects, uneven lighting, reflections, and backscatter, yielding robust models that perform effectively during real inspections \cite{alvarez2023mimir}. Similarly, OceanSim employs physics-based, GPU-accelerated rendering for cameras and sonar to rapidly generate realistic imagery \cite{song2025oceansim}. Additionally, converting ray-traced sonar data into lifelike images enhances acoustic perception for object detection \cite{sung2020realistic}.
In control, frameworks like Angler integrate Gazebo-based hydrodynamics simulation with ArduPilot software-in-loop control using Docker, ensuring the same code runs in simulation and on the real vehicle, achieving sub-meter station-keeping accuracy \cite{palmer2024angler}. For planning, Angler employs behavior trees for high-level decision-making, enabling the testing of complex intervention missions in simulation before deployment.

Despite these advances, detailed virtual environments still fall short of capturing all the nuances of the real underwater world. Incorporating deliberate disturbances such as realistic sensor inaccuracies, variable obstacle configurations, and fluctuating currents is essential to bridge the remaining gap and ensure robust field performance.

\section{Discussions}
\label{sec:discussion}
One of the ongoing difficulties with simulating marine conditions is achieving a balance between realism and computational efficiency. High-fidelity simulations are computationally intensive, which can limit their scalability and practicality for certain applications. Recent benchmarking frameworks \cite{huang2024urobench}, highlight the limitations in computational efficiency when deploying \ac{RL} algorithms in underwater simulators like HoloOcean, Project DAVE, and Stonefish. These tools demonstrate varied levels of resource utilization, which underscores the ongoing trade-offs between accuracy and performance. 
To address these challenges, optimizing hardware utilization is key. Simulators like MarineGym, which claims a 10,000-fold performance improvement on a single GPU \cite{chu2024marinegym}, leverage GPU acceleration for rendering and physics, enabling parallel processing for \ac{RL} training. This approach reduces training time, making it feasible for large-scale experiments, and could inspire similar optimizations in the state-of-the-art \acp{URS}. Effective sim-to-real transfer also requires accurate modeling of real-world physics, hydrodynamics, and sensor behaviors. HoloOcean’s realistic sonar and sensor simulations, Project DAVE’s dynamic bathymetry, and Stonefish’s advanced hydrodynamics all aim to minimize the sim-to-real gap. Validation with real-world data, as seen in An Open-Source Benchmark Simulator \cite{von2022open}, is crucial. Techniques like domain randomization and real data integration can further enhance transfer, ensuring \ac{RL} policies trained in simulation perform well in real underwater environments.

Another concern is the absence of a standardized simulator for \ac{RL} which significantly hampers reproducibility, benchmarking, and the cohesive advancement of robotic research \cite{dulac2020empirical, brockman2016openai}. Creating an open, standardized \ac{RL} simulator tailored specifically to underwater robotics would streamline research efforts, promote transparency, and enhance cross-validation of methodologies and results across the research community.

\section{Conclusions and Future Work}
\label{sec:conclusion}

This review reveals that while high-fidelity simulations are instrumental in accurately capturing underwater dynamics and sensor behaviors, they are inherently resource-intensive. This trade-off between simulation detail and computational efficiency remains a critical challenge, one that could be alleviated through advances such as GPU acceleration, which shows promise in reducing training time for complex algorithms like reinforcement learning.

Bridging the sim-to-real gap continues to be a significant hurdle. Although techniques such as enhanced sensor modeling, domain randomization, and the incorporation of real-world data have improved transferability, ensuring that simulated environments translate into reliable field performance still requires further validation and iterative refinement.

Moreover, established simulators, originally built on older frameworks (e.g., ROS1), risk obsolescence unless proactive migration and continual support are prioritized. This transition is essential for keeping simulation platforms relevant and aligned with emerging technologies.

Finally, the absence of a standardized simulation framework for underwater reinforcement learning hampers reproducibility and benchmarking across the research community. Establishing a common, open-source standard could streamline development efforts, foster transparency, and facilitate consistent cross-validation of methodologies. Together, these insights highlight the need for continued innovation to balance fidelity with efficiency, enhance real-world applicability, and build sustainable, standardized tools that can drive further advancements in underwater robotics.

\section*{Acknowledgment}
This work was supported in part by the Engineering and Physical Sciences Research Council (EPSRC) under Grant EP/Y000862/1.

\bibliographystyle{IEEEtran}
\bibliography{bibtex/references}

\begin{thebibliography}{10}
\providecommand{\url}[1]{#1}
\csname url@samestyle\endcsname
\providecommand{\newblock}{\relax}
\providecommand{\bibinfo}[2]{#2}
\providecommand{\BIBentrySTDinterwordspacing}{\spaceskip=0pt\relax}
\providecommand{\BIBentryALTinterwordstretchfactor}{4}
\providecommand{\BIBentryALTinterwordspacing}{\spaceskip=\fontdimen2\font plus
\BIBentryALTinterwordstretchfactor\fontdimen3\font minus \fontdimen4\font\relax}
\providecommand{\BIBforeignlanguage}[2]{{%
\expandafter\ifx\csname l@#1\endcsname\relax
\typeout{** WARNING: IEEEtran.bst: No hyphenation pattern has been}%
\typeout{** loaded for the language `#1'. Using the pattern for}%
\typeout{** the default language instead.}%
\else
\language=\csname l@#1\endcsname
\fi
#2}}
\providecommand{\BIBdecl}{\relax}
\BIBdecl

\bibitem{neira2021review}
J.~Neira, C.~Sequeiros, R.~Huamani, E.~Machaca, P.~Fonseca, and W.~Nina, ``Review on unmanned underwater robotics, structure designs, materials, sensors, actuators, and navigation control,'' \emph{Journal of Robotics}, vol. 2021, no.~1, p. 5542920, 2021.

\bibitem{christensen2022recent}
L.~Christensen, J.~de~Gea~Fern{\'a}ndez, M.~Hildebrandt, C.~E.~S. Koch, and B.~Wehbe, ``Recent advances in ai for navigation and control of underwater robots,'' \emph{Current Robotics Reports}, vol.~3, no.~4, pp. 165--175, 2022.

\bibitem{wang2022development}
J.~Wang, Z.~Wu, H.~Dong, M.~Tan, and J.~Yu, ``Development and control of underwater gliding robots: A review,'' \emph{IEEE/CAA Journal of Automatica Sinica}, vol.~9, no.~9, pp. 1543--1560, 2022.

\bibitem{matsebe2008review}
O.~Matsebe, C.~Kumile, and N.~Tlale, ``A review of virtual simulators for autonomous underwater vehicles (auvs),'' \emph{IFAC Proceedings Volumes}, vol.~41, no.~1, pp. 31--37, 2008.

\bibitem{cook2014survey}
D.~Cook, A.~Vardy, and R.~Lewis, ``A survey of auv and robot simulators for multi-vehicle operations,'' in \emph{2014 IEEE/OES Autonomous Underwater Vehicles (AUV)}.\hskip 1em plus 0.5em minus 0.4em\relax IEEE, 2014, pp. 1--8.

\bibitem{collins2021review}
J.~Collins, S.~Chand, A.~Vanderkop, and D.~Howard, ``A review of physics simulators for robotic applications,'' \emph{IEEE Access}, vol.~9, pp. 51\,416--51\,431, 2021.

\bibitem{ciuccoli2024underwater}
N.~Ciuccoli, L.~Screpanti, and D.~Scaradozzi, ``Underwater simulators analysis for digital twinning,'' \emph{IEEE Access}, 2024.

\bibitem{zhang2022dave}
M.~M. Zhang, W.-S. Choi, J.~Herman, D.~Davis, C.~Vogt, M.~McCarrin, Y.~Vijay, D.~Dutia, W.~Lew, S.~Peters \emph{et~al.}, ``Dave aquatic virtual environment: Toward a general underwater robotics simulator,'' in \emph{2022 IEEE/OES Autonomous Underwater Vehicles Symposium (AUV)}.\hskip 1em plus 0.5em minus 0.4em\relax IEEE, 2022, pp. 1--8.

\bibitem{amer2023unav}
A.~Amer, O.~{\'A}lvarez-Tu{\~n}{\'o}n, H.~{\.I}. U{\u{g}}urlu, J.~L.~F. Sejersen, Y.~Brodskiy, and E.~Kayacan, ``Unav-sim: A visually realistic underwater robotics simulator and synthetic data-generation framework,'' in \emph{2023 21st International Conference on Advanced Robotics (ICAR)}.\hskip 1em plus 0.5em minus 0.4em\relax IEEE, 2023, pp. 570--576.

\bibitem{cieslak2019stonefish}
P.~Cie{\'s}lak, ``Stonefish: An advanced open-source simulation tool designed for marine robotics, with a ros interface,'' in \emph{OCEANS 2019-Marseille}.\hskip 1em plus 0.5em minus 0.4em\relax IEEE, 2019, pp. 1--6.

\bibitem{pi2021twinbot}
R.~Pi, P.~Cie{\'s}lak, P.~Ridao, and P.~J. Sanz, ``Twinbot: Autonomous underwater cooperative transportation,'' \emph{IEEE Access}, vol.~9, pp. 37\,668--37\,684, 2021.

\bibitem{potokar2022holoocean}
E.~Potokar, S.~Ashford, M.~Kaess, and J.~G. Mangelson, ``Holoocean: An underwater robotics simulator,'' in \emph{2022 International Conference on Robotics and Automation (ICRA)}.\hskip 1em plus 0.5em minus 0.4em\relax IEEE, 2022, pp. 3040--3046.

\bibitem{dosovitskiy2017carla}
A.~Dosovitskiy, G.~Ros, F.~Codevilla, A.~Lopez, and V.~Koltun, ``Carla: An open urban driving simulator,'' in \emph{Conference on robot learning}.\hskip 1em plus 0.5em minus 0.4em\relax PMLR, 2017, pp. 1--16.

\bibitem{shah2018airsim}
S.~Shah, D.~Dey, C.~Lovett, and A.~Kapoor, ``Airsim: High-fidelity visual and physical simulation for autonomous vehicles,'' in \emph{Field and Service Robotics: Results of the 11th International Conference}.\hskip 1em plus 0.5em minus 0.4em\relax Springer, 2018, pp. 621--635.

\bibitem{todorov2012mujoco}
E.~Todorov, T.~Erez, and Y.~Tassa, ``Mujoco: A physics engine for model-based control,'' in \emph{2012 IEEE/RSJ international conference on intelligent robots and systems}.\hskip 1em plus 0.5em minus 0.4em\relax IEEE, 2012, pp. 5026--5033.

\bibitem{varshosaz2024monitoring}
M.~Varshosaz and A.~W{\k{a}}sowski, ``Monitoring safety and reliability of underwater robots: A case study,'' in \emph{International Conference on Bridging the Gap between AI and Reality}.\hskip 1em plus 0.5em minus 0.4em\relax Springer, 2024, pp. 291--299.

\bibitem{choi2021physics}
W.-S. Choi, D.~R. Olson, D.~Davis, M.~Zhang, A.~Racson, B.~Bingham, M.~McCarrin, C.~Vogt, and J.~Herman, ``Physics-based modelling and simulation of multibeam echosounder perception for autonomous underwater manipulation,'' \emph{Frontiers in Robotics and AI}, vol.~8, p. 706646, 2021.

\bibitem{gaspar2023feature}
A.~R. Gaspar and A.~Matos, ``Feature-based place recognition using forward-looking sonar,'' \emph{Journal of Marine Science and Engineering}, vol.~11, no.~11, p. 2198, 2023.

\bibitem{ruscio2023autonomous}
F.~Ruscio, R.~Costanzi, N.~Gracias, J.~Quintana, and R.~Garcia, ``Autonomous boundary inspection of posidonia oceanica meadows using an underwater robot,'' \emph{Ocean Engineering}, vol. 283, p. 114988, 2023.

\bibitem{gaspar2023visual}
A.~R. Gaspar, A.~Nunes, and A.~Matos, ``Visual place recognition for harbour infrastructures inspection,'' in \emph{OCEANS 2023-Limerick}.\hskip 1em plus 0.5em minus 0.4em\relax IEEE, 2023, pp. 1--9.

\bibitem{fossen1999guidance}
T.~I. Fossen, \emph{Guidance and control of ocean vehicles}, 1999.

\bibitem{deng2024simulating}
J.~Deng, A.~Filisetti, H.~S. Lim, D.~Y. Kim, and A.~Al-Hourani, ``Simulating sensor noise model for real-time testing in a virtual underwater environment,'' in \emph{2024 IEEE Annual Congress on Artificial Intelligence of Things (AIoT)}.\hskip 1em plus 0.5em minus 0.4em\relax IEEE, 2024, pp. 226--231.

\bibitem{hu2024disturbance}
Y.~Hu, B.~Li, B.~Jiang, J.~Han, and C.-Y. Wen, ``Disturbance observer-based model predictive control for an unmanned underwater vehicle,'' \emph{Journal of Marine Science and Engineering}, vol.~12, no.~1, p.~94, 2024.

\bibitem{song2025oceansim}
J.~Song, H.~Ma, O.~Bagoren, A.~V. Sethuraman, Y.~Zhang, and K.~A. Skinner, ``Oceansim: A gpu-accelerated underwater robot perception simulation framework,'' 2025.

\bibitem{akkaynak2018revised}
D.~Akkaynak and T.~Treibitz, ``A revised underwater image formation model,'' in \emph{2018 IEEE/CVF Conference on Computer Vision and Pattern Recognition}, 2018, pp. 6723--6732.

\bibitem{abdullah2024human}
A.~Abdullah, R.~Chen, D.~Blow, T.~Uthai, E.~J. Du, and M.~J. Islam, ``Human-machine interfaces for subsea telerobotics: From soda-straw to natural language interactions,'' \emph{arXiv preprint arXiv:2412.01753}, 2024.

\bibitem{dhurandher2008uwsim}
S.~K. Dhurandher, S.~Misra, M.~S. Obaidat, and S.~Khairwal, ``Uwsim: A simulator for underwater sensor networks,'' \emph{Simulation}, vol.~84, no.~7, pp. 327--338, 2008.

\bibitem{palomeras2012cola2}
N.~Palomeras, A.~El-Fakdi, M.~Carreras, and P.~Ridao, ``Cola2: A control architecture for auvs,'' \emph{IEEE Journal of Oceanic Engineering}, vol.~37, no.~4, pp. 695--716, 2012.

\bibitem{conti2017free}
R.~Conti, F.~Fanelli, E.~Meli, A.~Ridolfi, and R.~Costanzi, ``A free floating manipulation strategy for autonomous underwater vehicles,'' \emph{Robotics and Autonomous Systems}, vol.~87, pp. 133--146, 2017.

\bibitem{tosik2014mars}
T.~Tosik and E.~Maehle, ``Mars: A simulation environment for marine robotics,'' in \emph{2014 Oceans-St. John's}.\hskip 1em plus 0.5em minus 0.4em\relax IEEE, 2014, pp. 1--7.

\bibitem{manhaes2016uuv}
M.~M.~M. Manh{\~a}es, S.~A. Scherer, M.~Voss, L.~R. Douat, and T.~Rauschenbach, ``Uuv simulator: A gazebo-based package for underwater intervention and multi-robot simulation,'' in \emph{Oceans 2016 Mts/Ieee Monterey}.\hskip 1em plus 0.5em minus 0.4em\relax Ieee, 2016, pp. 1--8.

\bibitem{henriksen2016uw}
E.~H. Henriksen, I.~Schj{\o}lberg, and T.~B. Gjersvik, ``Uw morse: The underwater modular open robot simulation engine,'' in \emph{2016 IEEE/OES Autonomous Underwater Vehicles (AUV)}.\hskip 1em plus 0.5em minus 0.4em\relax IEEE, 2016, pp. 261--267.

\bibitem{garg2020netmarsys}
S.~Garg, J.~Quintas, J.~Cruz, and A.~M. Pascoal, ``Netmarsys-a tool for the simulation and visualization of distributed autonomous marine robotic systems,'' in \emph{2020 IEEE/OES Autonomous Underwater Vehicles Symposium (AUV)}.\hskip 1em plus 0.5em minus 0.4em\relax IEEE, 2020, pp. 1--5.

\bibitem{gezer2022working}
E.~C. Gezer, M.~Zhou, L.~Zhao, and W.~McConnell, ``Working toward the development of a generic marine vehicle framework: Ros-mvp,'' in \emph{OCEANS 2022, Hampton Roads}, 2022, pp. 1--5.

\bibitem{lonvcar2022marus}
I.~Lon{\v{c}}ar, J.~Obradovi{\'c}, N.~Kra{\v{s}}evac, L.~Mandi{\'c}, I.~Kvasi{\'c}, F.~Ferreira, V.~Slo{\v{s}}i{\'c}, {\DJ}.~Na{\dj}, and N.~Mi{\v{s}}kovi{\'c}, ``Marus-a marine robotics simulator,'' in \emph{OCEANS 2022, Hampton Roads}.\hskip 1em plus 0.5em minus 0.4em\relax IEEE, 2022, pp. 1--7.

\bibitem{lin2022oystersim}
X.~Lin, N.~Jha, M.~Joshi, N.~Karapetyan, Y.~Aloimonos, and M.~Yu, ``Oystersim: Underwater simulation for enhancing oyster reef monitoring,'' in \emph{OCEANS 2022, Hampton Roads}.\hskip 1em plus 0.5em minus 0.4em\relax IEEE, 2022, pp. 1--6.

\bibitem{liu2022fishgym}
W.~Liu, K.~Bai, X.~He, S.~Song, C.~Zheng, and X.~Liu, ``Fishgym: A high-performance physics-based simulation framework for underwater robot learning,'' in \emph{2022 International Conference on Robotics and Automation (ICRA)}.\hskip 1em plus 0.5em minus 0.4em\relax IEEE, 2022, pp. 6268--6275.

\bibitem{von2022open}
M.~von Benzon, F.~F. S{\o}rensen, E.~Uth, J.~Jouffroy, J.~Liniger, and S.~Pedersen, ``An open-source benchmark simulator: Control of a bluerov2 underwater robot,'' \emph{Journal of marine science and engineering}, vol.~10, no.~12, p. 1898, 2022.

\bibitem{palmer2024angler}
E.~Palmer, C.~Holm, and G.~Hollinger, ``Angler: An autonomy framework for intervention tasks with lightweight underwater vehicle manipulator systems,'' in \emph{2024 IEEE International Conference on Robotics and Automation (ICRA)}.\hskip 1em plus 0.5em minus 0.4em\relax IEEE, 2024, pp. 6126--6132.

\bibitem{huang2024urobench}
Z.~Huang, M.~Buchholz, M.~Grimaldi, H.~Yu, I.~Carlucho, and Y.~R. Petillot, ``Urobench: Comparative analyses of underwater robotics simulators from reinforcement learning perspective,'' in \emph{OCEANS 2024-Singapore}.\hskip 1em plus 0.5em minus 0.4em\relax IEEE, 2024, pp. 1--8.

\bibitem{marnet2024bridging}
L.~R. Marnet, S.~Grasshof, Y.~Brodskiy, and A.~W{\k{a}}sowski, ``Bridging the sim-to-real gap for underwater image segmentation,'' in \emph{OCEANS 2024-Singapore}.\hskip 1em plus 0.5em minus 0.4em\relax IEEE, 2024, pp. 1--6.

\bibitem{corsi2024aquatic}
D.~Corsi, D.~Camponogara, and A.~Farinelli, ``Aquatic navigation: A challenging benchmark for deep reinforcement learning,'' \emph{arXiv preprint arXiv:2405.20534}, 2024.

\bibitem{sethuraman2024machine}
A.~V. Sethuraman, A.~Sheppard, O.~Bagoren, C.~Pinnow, J.~Anderson, T.~C. Havens, and K.~A. Skinner, ``Machine learning for shipwreck segmentation from side scan sonar imagery: Dataset and benchmark,'' \emph{The International Journal of Robotics Research}, p. 02783649241266853, 2024.

\bibitem{watanabe2015rock}
T.~Watanabe, G.~Neves, R.~Cerqueira, T.~Trocoli, M.~Reis, S.~Joyeux, and J.~Albiez, ``The rock-gazebo integration and a real-time auv simulation,'' in \emph{2015 12th Latin American Robotics Symposium and 2015 3rd Brazilian Symposium on Robotics (LARS-SBR)}.\hskip 1em plus 0.5em minus 0.4em\relax IEEE, 2015, pp. 132--138.

\bibitem{katara2019open}
P.~Katara, M.~Khanna, H.~Nagar, and A.~Panaiyappan, ``Open source simulator for unmanned underwater vehicles using ros and unity3d,'' in \emph{2019 IEEE Underwater Technology (UT)}.\hskip 1em plus 0.5em minus 0.4em\relax IEEE, 2019, pp. 1--7.

\bibitem{palnitkar2023chatsim}
A.~Palnitkar, R.~Kapu, X.~Lin, C.~Liu, N.~Karapetyan, and Y.~Aloimonos, ``Chatsim: Underwater simulation with natural language prompting,'' in \emph{OCEANS 2023-MTS/IEEE US Gulf Coast}.\hskip 1em plus 0.5em minus 0.4em\relax IEEE, 2023, pp. 1--7.

\bibitem{chu2024marinegym}
S.~Chu, Z.~Huang, M.~Lin, D.~Li, and I.~Carlucho, ``Marinegym: Accelerated training for underwater vehicles with high-fidelity rl simulation,'' \emph{arXiv preprint arXiv:2410.14117}, 2024.

\bibitem{chaudhary2021development}
A.~Chaudhary, R.~Mishra, B.~Kalyan, and M.~Chitre, ``Development of an underwater simulator using unity3d and robot operating system,'' in \emph{OCEANS 2021: San Diego--Porto}.\hskip 1em plus 0.5em minus 0.4em\relax IEEE, 2021, pp. 1--7.

\bibitem{guerrero2021adaptive}
E.~Guerrero, F.~Bonin-Font, and G.~Oliver, ``Adaptive visual information gathering for autonomous exploration of underwater environments,'' \emph{IEEE Access}, vol.~9, pp. 136\,487--136\,506, 2021.

\bibitem{xie2024bathymetric}
Y.~Xie, G.~Troni, N.~Bore, and J.~Folkesson, ``Bathymetric surveying with imaging sonar using neural volume rendering,'' \emph{IEEE Robotics and Automation Letters}, vol.~9, no.~9, pp. 8146--8153, 2024.

\bibitem{zhou2022acrobatic}
M.~Zhou, E.~C. Gezer, W.~McConnell, and C.~Yuan, ``Acrobatic low-cost portable hybrid auv (alpha): System design and preliminary results,'' in \emph{OCEANS 2022, Hampton Roads}, 2022, pp. 1--5.

\bibitem{wozniak2024using}
G.~Wozniak, S.~Bhat, and I.~Stenius, ``Using reinforcement learning for hydrobatic maneuvering with autonomous underwater vehicles,'' in \emph{OCEANS 2024 - Singapore}, 2024, pp. 1--8.

\bibitem{munoz2024learning}
B.~Muñoz and G.~Troni, ``Learning the ego-motion of an underwater imaging sonar: A comparative experimental evaluation of novel cnn and rcnn approaches,'' \emph{IEEE Robotics and Automation Letters}, vol.~9, no.~3, pp. 2072--2079, 2024.

\bibitem{grimaldi2025stonefish}
M.~Grimaldi, P.~Cieslak, E.~Ochoa, V.~Bharti, H.~Rajani, I.~Carlucho, M.~Koskinopoulou, Y.~R. Petillot, and N.~Gracias, ``Stonefish: Supporting machine learning research in marine robotics,'' \emph{arXiv preprint arXiv:2502.11887}, 2025.

\bibitem{vaughnDSSim}
I.~Vaughn and S.~Suman, ``{Deep Submergence ds\_sim},'' \url{https://bitbucket.org/whoidsl/ds_sim/src/master/}, accessed: 2025-03-18.

\bibitem{voskakis2024modeling}
D.~Voskakis, E.~Kelasidi, and N.~Papandroulakis, ``Modeling and control of an underwater calibration arm,'' in \emph{2024 32nd Mediterranean Conference on Control and Automation (MED)}, 2024, pp. 81--87.

\bibitem{lensgraf2024scalable}
S.~Lensgraf, A.~Sarkar, A.~Pediredla, D.~Balkcom, and A.~Q. Li, ``Scalable underwater assembly with reconfigurable visual fiducials,'' in \emph{2024 IEEE International Conference on Robotics and Automation (ICRA)}, 2024, pp. 3639--3645.

\bibitem{pan2025russorobustunderwaterslam}
\BIBentryALTinterwordspacing
S.~Pan, Z.~Hong, Z.~Hu, X.~Xu, W.~Lu, and L.~Hu, ``Russo: Robust underwater slam with sonar optimization against visual degradation,'' 2025. [Online]. Available: \url{https://arxiv.org/abs/2503.01434}
\BIBentrySTDinterwordspacing

\bibitem{chan2023predicting}
W.~Y. Chan, M.~Ludvigsen, and E.~Kelasidi, ``Predicting underwater vehicle manipulator system dynamics using time-series methods,'' in \emph{OCEANS 2023 - MTS/IEEE U.S. Gulf Coast}, 2023, pp. 1--5.

\bibitem{mane2024eroas3defficientreactive}
\BIBentryALTinterwordspacing
P.~Mane, A.~J. George, R.~Makam, R.~Majumder, and S.~Sundaram, ``Eroas: 3d efficient reactive obstacle avoidance system for autonomous underwater vehicles using 2.5d forward-looking sonar,'' 2024. [Online]. Available: \url{https://arxiv.org/abs/2411.05516}
\BIBentrySTDinterwordspacing

\bibitem{potokar2024holoocean}
E.~Potokar, K.~Lay, K.~Norman, D.~Benham, S.~Ashford, R.~Peirce, T.~B. Neilsen, M.~Kaess, and J.~G. Mangelson, ``Holoocean: A full-featured marine robotics simulator for perception and autonomy,'' \emph{IEEE Journal of Oceanic Engineering}, 2024.

\bibitem{yang2023oceanchat}
R.~Yang, M.~Hou, J.~Wang, and F.~Zhang, ``Oceanchat: Piloting autonomous underwater vehicles in natural language,'' \emph{arXiv preprint arXiv:2309.16052}, 2023.

\bibitem{qadri2023neural}
M.~Qadri, M.~Kaess, and I.~Gkioulekas, ``Neural implicit surface reconstruction using imaging sonar,'' in \emph{2023 IEEE International Conference on Robotics and Automation (ICRA)}.\hskip 1em plus 0.5em minus 0.4em\relax IEEE, 2023, pp. 1040--1047.

\bibitem{norman2023actag}
K.~Norman, D.~Butterfield, and J.~G. Mangelson, ``Actag: Opti-acoustic fiducial markers for underwater localization and mapping,'' in \emph{2023 IEEE/RSJ International Conference on Intelligent Robots and Systems (IROS)}.\hskip 1em plus 0.5em minus 0.4em\relax IEEE, 2023, pp. 9955--9962.

\bibitem{wang2024invariant}
C.~Wang, C.~Cheng, C.~Cao, X.~Guo, G.~Pan, and F.~Zhang, ``An invariant filtering method based on frame transformed for underwater ins/dvl/ps navigation,'' \emph{Journal of Marine Science and Engineering}, vol.~12, no.~7, p. 1178, 2024.

\bibitem{lennox2024collaborative}
B.~Lennox, K.~Groves \emph{et~al.}, ``Collaborative aquatic positioning system utilising multi-beam sonar and depth sensors,'' \emph{arXiv preprint arXiv:2403.10397}, 2024.

\bibitem{nadj2022towards}
{\DJ}.~Na{\dj}, F.~Ferreira, I.~Kvasi{\'c}, L.~Mandi{\'c}, V.~Slo{\v{s}}i{\'c}, C.~Walker, D.~O. Antillon, and I.~Anderson, ``Towards robot-aided diver navigation in mapped environments (roadmap),'' in \emph{OCEANS 2022, Hampton Roads}.\hskip 1em plus 0.5em minus 0.4em\relax IEEE, 2022, pp. 1--5.

\bibitem{campos2021orb}
C.~Campos, R.~Elvira, J.~J.~G. Rodr{\'\i}guez, J.~M. Montiel, and J.~D. Tard{\'o}s, ``Orb-slam3: An accurate open-source library for visual, visual--inertial, and multimap slam,'' \emph{IEEE transactions on robotics}, vol.~37, no.~6, pp. 1874--1890, 2021.

\bibitem{wang2021tartanvo}
W.~Wang, Y.~Hu, and S.~Scherer, ``Tartanvo: A generalizable learning-based vo,'' in \emph{Conference on Robot Learning}.\hskip 1em plus 0.5em minus 0.4em\relax PMLR, 2021, pp. 1761--1772.

\bibitem{aubard2024mission}
M.~Aubard, S.~Quijano, O.~Álvarez Tuñón, L.~Antal, M.~Costa, and Y.~Brodskiy, ``Mission planning and safety assessment for pipeline inspection using autonomous underwater vehicles: A framework based on behavior trees,'' in \emph{OCEANS 2024 - Singapore}, 2024, pp. 1--6.

\bibitem{alvarez2023mimir}
O.~{\'A}lvarez-Tu{\~n}{\'o}n, H.~Kanner, L.~R. Marnet, H.~X. Pham, J.~le~Fevre~Sejersen, Y.~Brodskiy, and E.~Kayacan, ``Mimir-uw: A multipurpose synthetic dataset for underwater navigation and inspection,'' in \emph{2023 IEEE/RSJ International Conference on Intelligent Robots and Systems (IROS)}.\hskip 1em plus 0.5em minus 0.4em\relax IEEE, 2023, pp. 6141--6148.

\bibitem{sung2020realistic}
M.~Sung, J.~Kim, M.~Lee, B.~Kim, T.~Kim, J.~Kim, and S.-C. Yu, ``Realistic sonar image simulation using deep learning for underwater object detection,'' \emph{International Journal of Control, Automation and Systems}, vol.~18, no.~3, pp. 523--534, 2020.

\bibitem{dulac2020empirical}
G.~Dulac-Arnold, N.~Levine, D.~J. Mankowitz, J.~Li, C.~Paduraru, S.~Gowal, and T.~Hester, ``An empirical investigation of the challenges of real-world reinforcement learning,'' \emph{arXiv preprint arXiv:2003.11881}, 2020.

\bibitem{brockman2016openai}
G.~Brockman, V.~Cheung, L.~Pettersson, J.~Schneider, J.~Schulman, J.~Tang, and W.~Zaremba, ``Openai gym,'' \emph{arXiv preprint arXiv:1606.01540}, 2016.

\end{thebibliography}

\end{document}